\documentclass[10pt,twocolumn,letterpaper]{article}

\makeatletter
\@namedef{ver@eso-pic.sty}{}
\makeatother

\let\LenToUnit\relax

\usepackage{iccv}

\usepackage{float}
\usepackage{pdfpages}
\usepackage[absolute,overlay]{textpos}
\usepackage{placeins}

\AtBeginDocument{
  \hypersetup{pageanchor=false}
}

\definecolor{iccvblue}{rgb}{0.21,0.49,0.74}
\usepackage[pagebackref,breaklinks,colorlinks,allcolors=iccvblue]{hyperref}

\def\paperID{12178}
\def\confName{ICCV}
\def\confYear{2025}

\title{Attention, Please! PixelSHAP Reveals What Vision-Language Models Actually Focus On}

\author{
Roni Goldshmidt\\
Nexar, Tel Aviv\\
{\tt\small roni.goldshmidt@getnexar.com}
}

\begin{document}
\maketitle

\pagenumbering{gobble}

% \ificcvfinal\thispagestyle{empty}\fi

% Sections
\section*{Abstract}
\emph{
Interpretability in Vision-Language Models (VLMs) is essential for building trust, enabling debugging, and supporting critical decision-making in high-stakes applications. We introduce \textbf{PixelSHAP}, a powerful model-agnostic framework that extends Shapley-based analysis to structured visual entities. While TokenSHAP \cite{tokenShap2023} successfully applies to text prompts, PixelSHAP advances this approach for vision-based reasoning by systematically analyzing image objects and precisely quantifying their influence on a VLM's response.
}
\emph{
PixelSHAP operates solely on input-output pairs without requiring access to model internals, ensuring seamless compatibility with both open-source and commercial models. The framework supports diverse embedding-based similarity metrics and achieves remarkable computational efficiency through innovative optimization techniques derived from TokenSHAP \cite{tokenShap2023}.
}
\emph{
We demonstrate PixelSHAP's exceptional ability to enhance interpretability of complex visual scenes, with potential applications in critical domains such as autonomous driving. Our open-source implementation provides researchers and practitioners with a robust tool for advancing explainable AI in vision-language systems.
}    
\section{Introduction}

Interpretability in Vision-Language Models (VLMs) is essential for trust, debugging, and decision-making, particularly in high-stakes applications such as autonomous driving. Existing techniques primarily focus on token-level or feature-level analysis \cite{tokenShap2023, lime2016, ribeiro2018anchors}, but fail to capture \textbf{spatial object relevance}, leaving it unclear which specific objects influenced a model's response. For example, if a VLM generates the caption \textit{"A pedestrian is crossing the street"} for an image with multiple pedestrians, traditional methods cannot identify which specific pedestrian the model focused on.

To address this limitation, we introduce \textbf{PixelSHAP}, the first object-level interpretability framework for generative VLMs that extends Shapley-based analysis to structured visual entities. Unlike previous approaches that work only with classification models or require access to model internals, PixelSHAP operates solely on input-output pairs from models that generate open-ended text responses. While TokenSHAP \cite{tokenShap2023} successfully applies Shapley value estimation to textual prompts, PixelSHAP adapts this concept to visual content by defining image objects as structured \textbf{feature groups}, systematically perturbing them, and quantifying their contribution to a VLM's response.

\begin{figure}[H]
    \centering
    \includegraphics[width=0.95\linewidth]{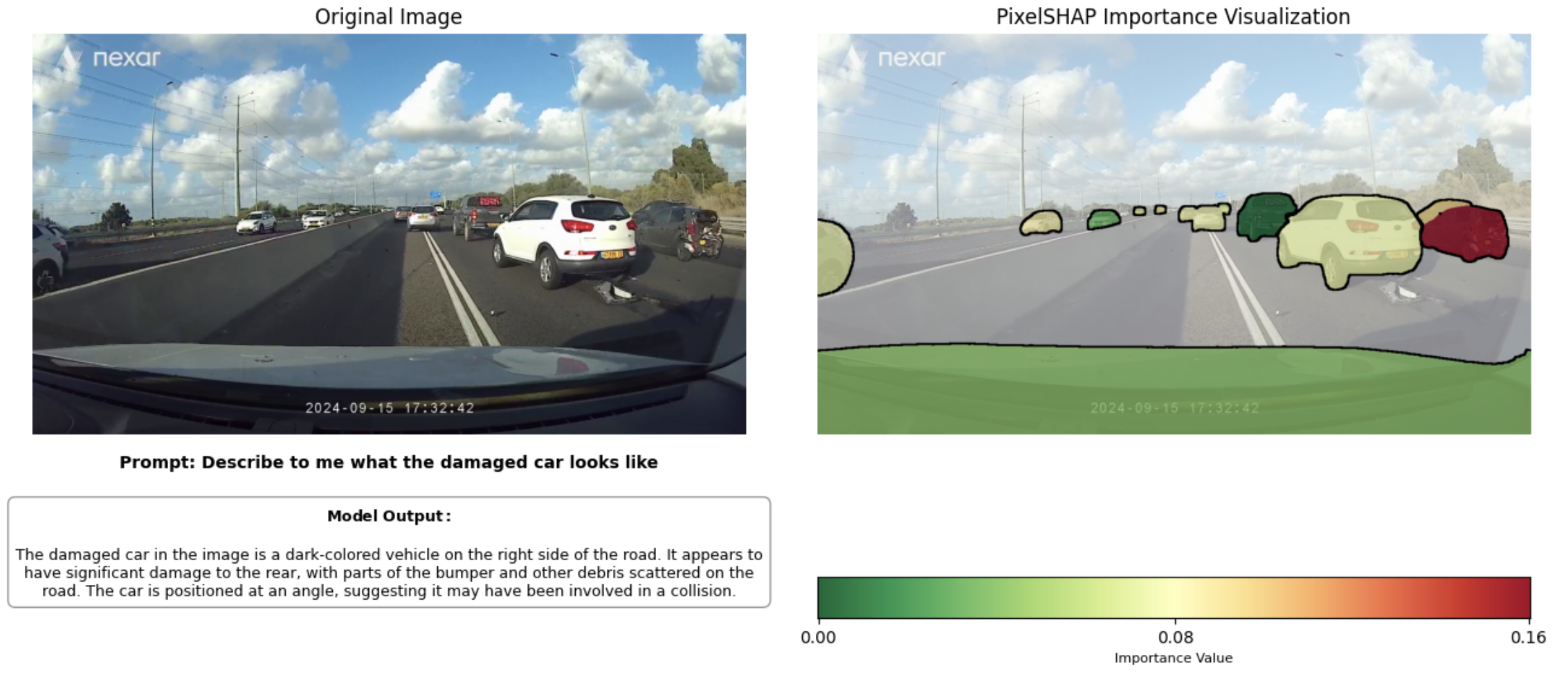}
    \caption{Pixel SHAP simulation of a car crash. The left shows the scene, prompt, and GPT-4o response, and the right highlights object importance.}
    \label{fig:pixelshap_example}
\end{figure}

PixelSHAP significantly improves computational efficiency by shifting from pixel-level to object-level perturbations. By leveraging segmentation models to generate object masks and applying Shapley-based importance estimation, our approach evaluates only a limited number of \textbf{meaningful objects} rather than thousands of individual pixels, making it computationally tractable while preserving high interpretability fidelity.

\subsection*{Contributions}
Our work introduces PixelSHAP and makes the following key contributions:
\begin{itemize}[leftmargin=*]
    \item \textbf{Object-Level Explainability for VLMs:} We propose PixelSHAP, the first model-agnostic interpretability framework for text-generative VLMs, integrating segmentation-based perturbations with Shapley value estimation.
    \item \textbf{Visual Attribution:} PixelSHAP provides a method for visualizing object contributions, allowing users to analyze model decisions through heatmaps and object masks.
    \item \textbf{Improved Trustworthiness:} By identifying which objects influenced a model's output, PixelSHAP enhances transparency in vision-language decision-making.
    \item \textbf{Challenges and Trade-offs:} We discuss the computational complexity and sensitivity to segmentation quality associated with object-level perturbation methods.
\end{itemize}

The remainder of this paper is structured as follows: Section~\ref{sec:related_work} reviews existing interpretability methods. Section~\ref{sec:problem_statement} details the problem statement. Section~\ref{sec:methodology} explains the PixelSHAP algorithm. Section~\ref{sec:dataset_and_masking} describes our dataset and masking strategies. Section~\ref{sec:results} presents empirical evaluations, Section~\ref{sec:qualitative} shows qualitative examples, Section~\ref{sec:discussion} discusses implications and future directions, and Section~\ref{sec:conclusion} concludes our work.
\section{Related Work}
\label{sec:related_work}

Interpretability in vision-language models (VLMs) has been studied through various approaches including perturbation-based analysis, gradient-based saliency maps, and Shapley-value estimations. We categorize existing methods into general interpretability, Shapley-based explanations, and vision-specific approaches.

\subsection{General Interpretability Methods}
Feature importance methods like LIME~\cite{lime2016} and Anchors~\cite{ribeiro2018anchors} provide locally interpretable explanations by perturbing inputs, while Grad-CAM~\cite{selvaraju2017grad} visualizes class-specific saliency maps for convolutional networks. More advanced techniques include RISE~\cite{petsiuk2018rise}, which uses randomized input sampling, and XRAI~\cite{sundararajan2019xrai}, which groups important pixels into meaningful regions. However, these methods typically don't account for structured objects within images.

\subsection{Shapley-Based Interpretability}
Shapley values from cooperative game theory~\cite{shapley1953} have become valuable for machine learning explainability due to their axiomatic properties. TokenSHAP~\cite{tokenShap2023} applied this approach to language models, quantifying individual token contributions, but doesn't extend naturally to vision content. FastSHAP~\cite{jethani2021fastshap} made Shapley estimation computationally feasible for large models, while MM-SHAP~\cite{parcalabescu2022mmshap} focused on attributing importance between visual and textual modalities using image patches rather than semantically meaningful objects, limiting its utility for visual interpretability.

\subsection{Interpretability in Vision-Language Models}
VLM interpretability remains challenging due to complex multimodal interactions. Methods like Score-CAM~\cite{wang2020scorecam} and CLIP-Explainer~\cite{chefer2021transformer} generate attention maps to visualize text-image associations. Newer approaches leverage GroundingDINO~\cite{liu2023groundingdino} and SAM~\cite{kirillov2023segment} for object recognition and segmentation. For black-box commercial VLMs, occlusion-based techniques~\cite{zeiler2014visualizing} and gradient-free approaches like D-RISE~\cite{petsiuk2021drise} and Meaningful Perturbation~\cite{fong2017interpretable} have been adapted to identify relevant image regions.

\subsection{Object-Level Explainability and Our Contribution}
Despite progress in interpretability, methods specifically designed for object-level explainability in text-generative VLMs are lacking. The increasing use of segmentation-based explainability~\cite{fong2019understanding} and multimodal reasoning frameworks~\cite{radford2021learning} highlights this need. While XAtten~\cite{gokhale2020xatten} proposed a cross-modal attention-based interpretability framework, it remains model-dependent, requiring access to model internals.

We introduce \textit{PixelSHAP}, extending TokenSHAP~\cite{tokenShap2023} to enable object-centric attribution using segmentation models. By systematically masking objects and evaluating their influence on VLM-generated text, PixelSHAP provides a structured understanding of vision-language interactions without requiring access to model internals. Unlike gradient-based methods that rely on activation maps, our approach directly quantifies object importance through systematic perturbation of visual entities, making it the first method to integrate segmentation-based object attribution with Shapley value estimation for text-generative VLMs.
\section{Problem Statement}
\label{sec:problem_statement}

\begin{figure*}[t]
    \centering
    \includegraphics[width=\textwidth]{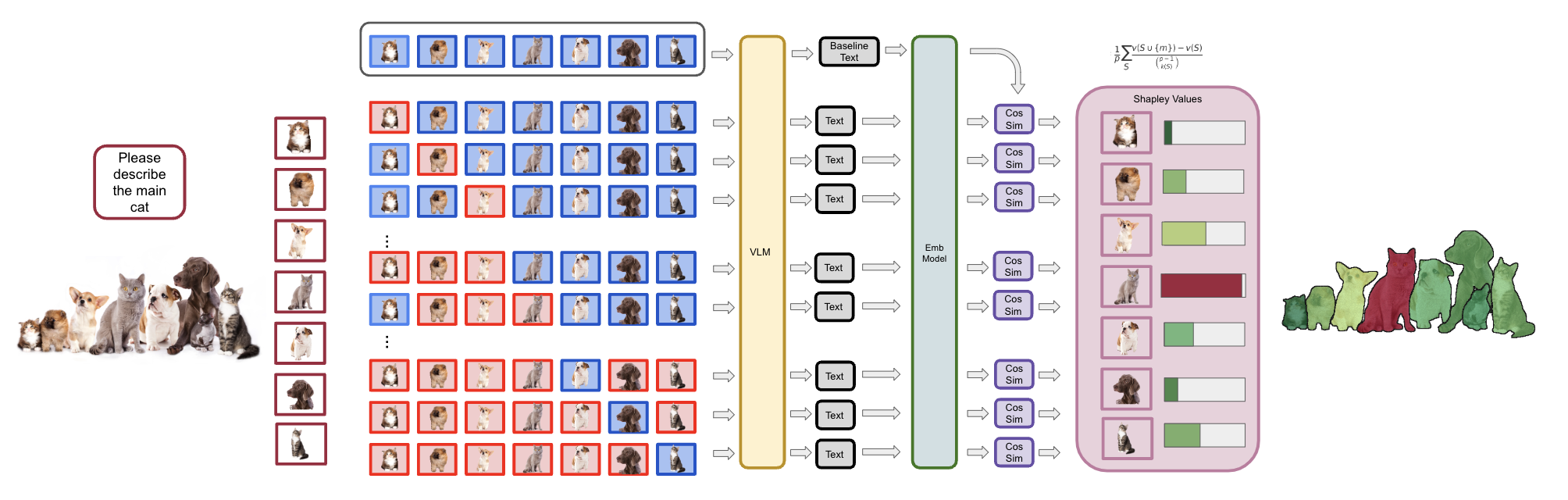}
    \caption{Overview of the PixelSHAP framework. The method systematically perturbs object groups, queries a vision-language model (VLM), and computes Shapley values to quantify object importance.}
    \label{fig:PixelSHAP_intro}
\end{figure*}

The core challenge in interpreting Vision-Language Models (VLMs) that generate free-form text lies in isolating the contribution of specific visual elements to the model's output. Unlike classification models that produce discrete probabilities over predefined categories, generative VLMs produce natural language that lacks a direct quantitative link to input features. This introduces a fundamental attribution problem: how do we determine which parts of an image most influenced a particular textual response?

Formally, given an image $I$ containing multiple objects $\{O_1, O_2, ..., O_n\}$, a text prompt $p$, and a VLM response $R = f(I, p)$, our goal is to attribute a relative importance score $\phi_i$ to each object $O_i$ that accurately reflects its influence on $R$. This attribution problem is challenging for several reasons:

\begin{enumerate}
    \item \textbf{Non-Discrete Outputs:} Unlike classification models that output probability distributions, text-generative VLMs produce sequences where influence cannot be directly mapped to class probabilities.
    
    \item \textbf{Semantic Ambiguity:} The influence of visual elements on text generation often has semantic dimensions that pixel-level perturbations cannot capture.
    
    \item \textbf{Black-Box Constraints:} Many commercial VLMs provide no access to internal representations, gradients, or attention weights.
    
    \item \textbf{Computational Complexity:} Naive pixel-level perturbation approaches would require an infeasible number of model queries for high-resolution images.
\end{enumerate}

While existing methods like RISE~\cite{petsiuk2018rise} and XRAI~\cite{sundararajan2019xrai} attempt to identify important image regions, they frequently blend multiple semantic entities into single importance regions. Consider an autonomous driving scenario where a VLM generates the caption \textit{"A pedestrian is crossing the street"}. Current methods cannot distinguish which specific pedestrian (if multiple are present) influenced this response, creating ambiguity in critical decision contexts.

PixelSHAP addresses these challenges by reformulating the attribution problem around objects rather than pixels. By leveraging semantic segmentation to define meaningful visual entities and applying Shapley-based importance estimation, our approach provides structured object-level attributions for any text-generating VLM without requiring access to model internals.
\section{Methodology}
\label{sec:methodology}

PixelSHAP adapts Shapley value analysis to visual objects, extending TokenSHAP's~\cite{tokenShap2023} approach from text to structured image regions. Our method operates in four key steps, as illustrated in Figure~\ref{fig:PixelSHAP_intro}.

\subsection{Object Detection and Segmentation}
Given an input image $\mathbf{I}$, we employ segmentation models to identify distinct visual entities:
\begin{equation}
    \{ \mathcal{O}_1, \mathcal{O}_2, ..., \mathcal{O}_N \} = f_{\text{seg}}(\mathbf{I})
\end{equation}
where $\mathcal{O}_i$ represents the $i^{th}$ object's segmentation mask. Our implementation uses GroundingDINO~\cite{liu2023groundingdino} for object detection followed by SAM~\cite{kirillov2023segment} for precise segmentation, though any segmentation approach is compatible with our framework and a customized model can be sent according to the use case.

\subsection{Group Formation and Object Representation}
We represent detected objects as a set of label-mask pairs:
\begin{equation}
    S = \{ (l_1, m_1), (l_2, m_2), ..., (l_N, m_N) \}
\end{equation}
where $l_i$ is the object label and $m_i$ is its corresponding segmentation mask. This approach allows PixelSHAP to adapt to different segmentation paradigms, from instance-level to semantic groupings.

\subsection{Selective Object Masking}
To quantify object importance, we systematically mask different object combinations and analyze their impact on the VLM's response. For a subset of objects $S^\prime \subseteq S$, we generate a modified image:
\begin{equation}
    \mathbf{I}^\prime = \mathcal{M}(S^\prime, \mathbf{I})
\end{equation}
where $\mathcal{M}$ is a masking function that removes objects not in $S^\prime$. For computational efficiency, we begin with first-order perturbations (masking each object independently), which alone achieve 92\% correlation with exact Shapley values ~\cite{tokenShap2023}.  We then enhance precision through strategic Monte Carlo sampling of object combinations.

The VLM generates a response to each modified image:
\begin{equation}
    y^\prime = f_{\text{VLM}}(\mathbf{I}^\prime, p)
\end{equation}
where $p$ is the original text prompt. 

\subsection{Shapley Value Computation}
The contribution of each object $i$ is computed using the standard Shapley formulation:
\begin{equation}
    \phi_i = \sum_{S^\prime \subseteq S \setminus \{i\}} \frac{|S^\prime|! (|S| - |S^\prime| - 1)!}{|S|!} \left[ v(S^\prime \cup \{i\}) - v(S^\prime) \right]
\end{equation}
where $v(S^\prime)$ measures the similarity between the VLM's responses to the original and modified images. We compute this using cosine similarity between text embeddings:
\begin{equation}
    v(S^\prime) = \cos(E(y), E(y^\prime))
\end{equation}
where $E(\cdot)$ is a text embedding function. Our experiments with GPT-4o used OpenAI's text-embedding-3-small, but any suitable embedding model can be employed.

The final Shapley values $\{\phi_1, \phi_2, ..., \phi_N\}$ provide a principled quantification of each object's contribution to the VLM's response, enabling fine-grained attribution without requiring access to the model's internal representations. This makes PixelSHAP applicable to any black-box VLM that accepts image inputs and produces text outputs.
\section{Dataset and Evaluation Framework}
\label{sec:dataset_and_masking}

\subsection{Object Importance Evaluation Dataset}
Most interpretability methods for computer vision are designed for classification models that produce probability distributions over predefined classes. However, VLMs generate open-ended textual responses rather than class probabilities, making traditional attribution evaluation methods unsuitable. To address this challenge, we constructed a specialized evaluation dataset focused on object-level attribution.

Our Object Importance Evaluation Dataset is built on the COCO validation set and designed specifically to benchmark how well different methods can identify the most relevant objects in an image given a contextual question. For each image, we presented GPT-4o with both the image and its corresponding COCO bounding box annotations, then prompted it to:
\begin{enumerate}
    \item Generate a focused natural language question requiring attention to a specific object
    \item Identify the target object that should be the focus when answering that question
\end{enumerate}

This process created a ground-truth dataset where each entry includes:
\begin{itemize}
    \item The image
    \item A list of all tagged objects and their bounding boxes
    \item The specific target object (with name and bounding box) that should be attended to
    \item The natural language question itself
\end{itemize}

\begin{figure}[H]
    \centering
    \includegraphics[width=\columnwidth,keepaspectratio]{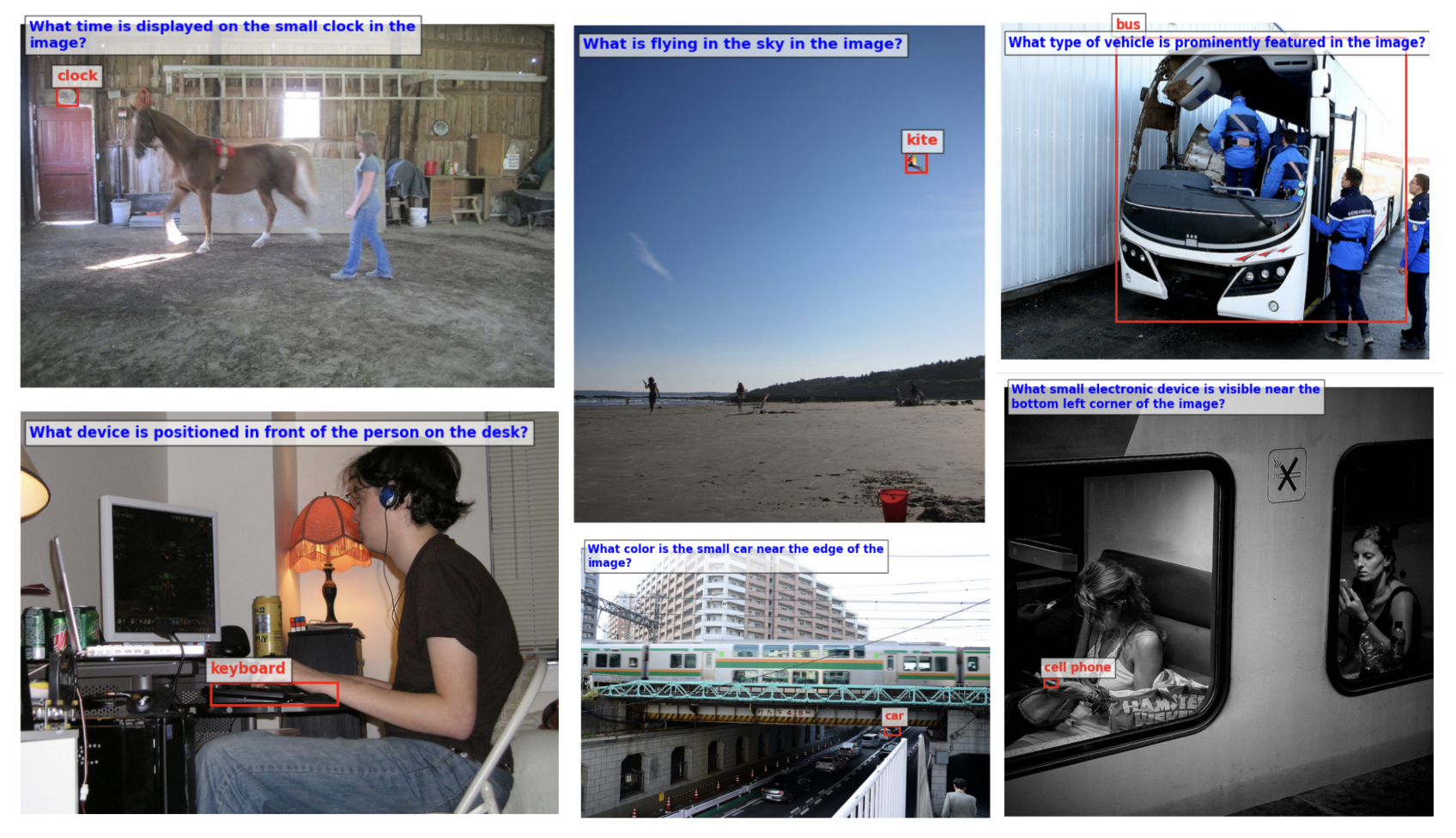}
    \caption{Sample images from our dataset, each with a question and a bounding box highlighting the object needed to answer it.}
    \label{fig:coco_vlm_example}
\end{figure}

The dataset allows us to quantitatively evaluate whether explainability methods can correctly identify which objects a VLM needed to focus on to generate its response. This dataset will be made publicly available upon publication.

\subsection{Masking Strategies and Their Rationale}
\label{subsec:masking_strategies}

A critical component of perturbation-based methods like PixelSHAP is how objects are masked when assessing their importance. We evaluated three distinct masking strategies, each addressing different challenges in object occlusion:

\begin{figure}[H]
    \centering
    \includegraphics[width=\columnwidth]{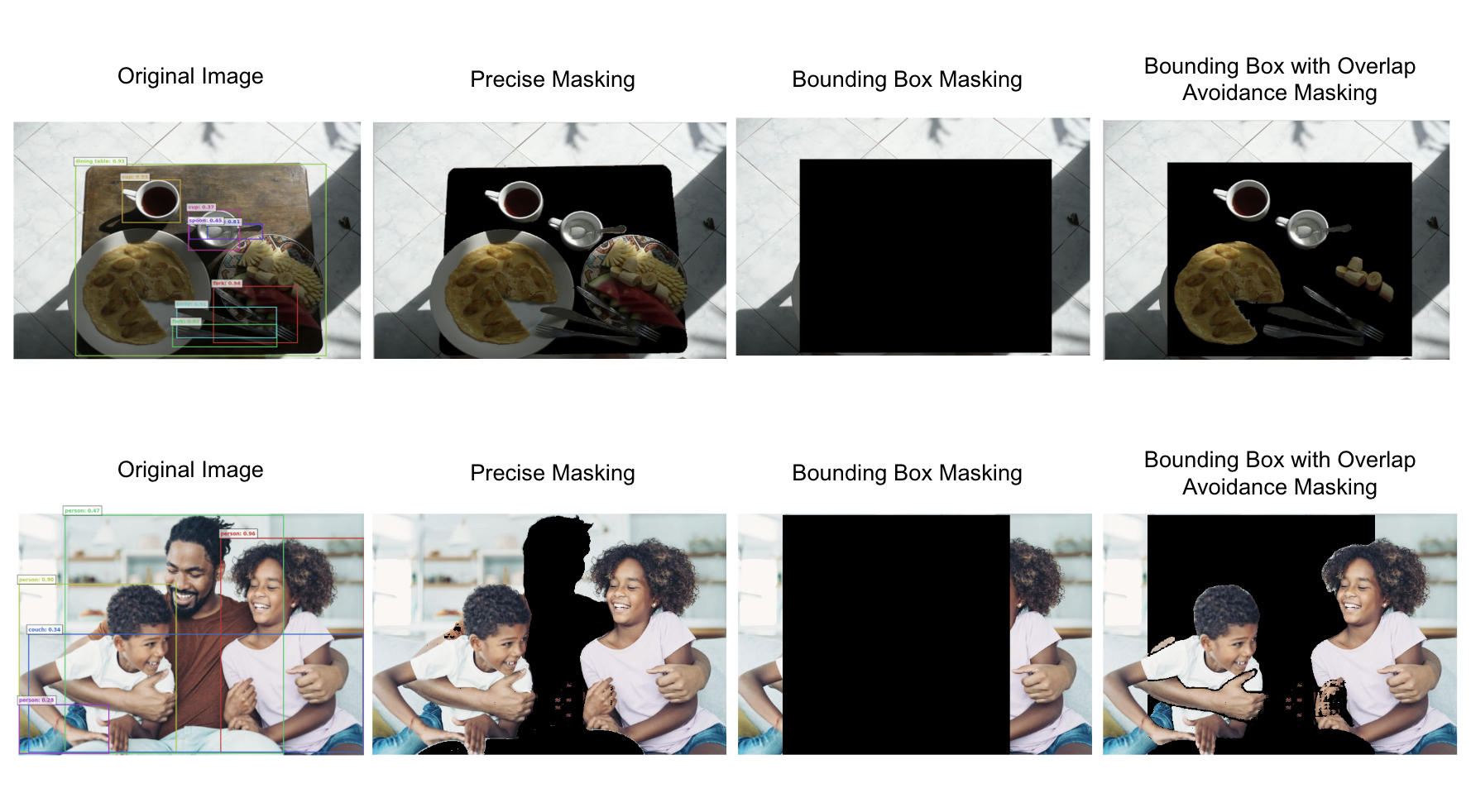}
    \caption{Comparison of three masking strategies: (a) Precise Masking follows object contours exactly but may leave recognizable silhouettes; (b) Bounding Box Masking completely occludes the object but may inadvertently mask neighboring objects; (c) Bounding Box with Overlap Avoidance (BBOA) masks the object while preserving the contours of adjacent objects.}
    \label{fig:masking_strategies}
\end{figure}

\begin{itemize}
    \item \textbf{Precise Masking}: Uses pixel-perfect instance segmentation masks conforming exactly to the object's shape. While this approach preserves surrounding context, it creates a silhouette that often leaves enough visual information for the model to recognize the masked object, undermining the perturbation's effectiveness.
    
    \item \textbf{Bounding Box Masking}: Uses rectangular bounding boxes that completely occlude the target object, preventing the model from inferring its presence from silhouettes. However, this approach often inadvertently masks portions of adjacent objects, introducing confounding variables when measuring the specific object's importance.
    
    \item \textbf{Bounding Box with Overlap Avoidance (BBOA)}: Our proposed hybrid approach that balances complete object occlusion with minimal interference to neighboring objects. BBOA uses a bounding box to mask the target object but selectively reveals the contours of other objects that would otherwise be occluded by the box. This preserves the segmentation integrity of non-target objects while ensuring the target object is completely hidden from the model.
\end{itemize}

BBOA addresses a fundamental tension in perturbation-based attribution: complete occlusion versus isolated impact. By hiding the target object completely (preventing recognition from silhouettes) while preserving adjacent objects' visibility, it allows for more accurate attribution of importance to specific objects rather than to regions of the image that may contain multiple entities.

\subsection{Experimental Setup}
To assess PixelSHAP's effectiveness and robustness, we evaluated it against two simple heuristic baselines:
\begin{itemize}
    \item \textbf{Largest Object}: Selecting the largest detected object based on bounding box area
    \item \textbf{Central Object}: Selecting the object closest to the image center
\end{itemize}

We conducted experiments across multiple vision-language models with varying architectures and capabilities:
\begin{itemize}
    \item \textbf{GPT-4o} \cite{chatgpt4o}: A commercial multimodal model from OpenAI
    \item \textbf{Gemini-2.0-flash} \cite{gemini2pro}: Google's multimodal foundation model
    \item \textbf{LLaVA-v1.5-7B} \cite{liu2023llava}: An open-source instruction-tuned VLM
    \item \textbf{LLaMA-3.2-11B-Vision} \cite{llama3.2}: Meta's open-source vision-language model
\end{itemize}

\subsection{Evaluation Metrics}
We employed the following metrics to quantitatively assess performance:
\begin{itemize}
    \item \textbf{Recall@K}: Measures whether the ground truth focus object appears in the top-$K$ most important objects identified by the method
    \item \textbf{IoU} (Intersection over Union): Spatial overlap between the predicted important region and the ground truth object region
    \item \textbf{Similarity Drop}: Change in model output similarity when the predicted important object is masked (larger drops indicate higher object importance)
\end{itemize}
\section{Experimental Results}
\label{sec:results}

We evaluated PixelSHAP across different VLMs and masking strategies using our Object Importance Evaluation Dataset. Table \ref{tab:accuracy_results} presents comprehensive results comparing performance across models, masking strategies, and baseline methods.

\begin{table}[!htbp]
\centering
\resizebox{\columnwidth}{!}{
\begin{tabular}{lccccccc}
\toprule
\textbf{Model} & \textbf{Masking} & \textbf{Comp. Time (s)} $\downarrow$ & \textbf{Sim. Drop} $\uparrow$ & \textbf{IoU@1} (\%) $\uparrow$ & \textbf{Recall@1} (\%) $\uparrow$ & \textbf{Recall@2} (\%) $\uparrow$ & \textbf{Recall@3} (\%) $\uparrow$ \\
\midrule
GPT-4o & \textbf{BBOA (ours)} & 52.95 $\pm$ 0.36 & 27.58 $\pm$ 0.19 & \textbf{47.12 $\pm$ 0.41} & \textbf{60.56 $\pm$ 0.39} & \textbf{77.31 $\pm$ 0.21} & \textbf{87.66 $\pm$ 0.28} \\
GPT-4o & precise & \textbf{52.22 $\pm$ 0.27} & 24.76 $\pm$ 0.14 & 44.61 $\pm$ 0.53 & 57.61 $\pm$ 0.81 & 76.36 $\pm$ 0.30 & 84.71 $\pm$ 0.25 \\
GPT-4o & bbox & 54.42 $\pm$ 0.38 & \textbf{30.26 $\pm$ 0.15} & 40.61 $\pm$ 0.32 & 53.18 $\pm$ 0.38 & 74.39 $\pm$ 0.31 & 85.20 $\pm$ 0.15 \\
\midrule
Gemini-2.0-flash & \textbf{BBOA (ours)} & 28.18 $\pm$ 0.13 & 25.60 $\pm$ 0.13 & \textbf{51.71 $\pm$ 1.08} & \textbf{67.48 $\pm$ 0.29} & \textbf{81.78 $\pm$ 0.27} & \textbf{89.17 $\pm$ 0.22} \\
Gemini-2.0-flash & precise & 27.93 $\pm$ 1.18 & 22.74 $\pm$ 0.27 & 45.61 $\pm$ 0.52 & 59.62 $\pm$ 0.21 & 76.34 $\pm$ 0.82 & 84.73 $\pm$ 0.26 \\
Gemini-2.0-flash & bbox & \textbf{27.81 $\pm$ 0.09} & \textbf{29.14 $\pm$ 0.15} & 42.90 $\pm$ 0.31 & 58.10 $\pm$ 0.45 & 79.83 $\pm$ 0.51 & 88.68 $\pm$ 0.23 \\
\midrule
LLaVA-v1.5-7B & \textbf{BBOA (ours)} & 17.28 $\pm$ 0.07 & 21.37 $\pm$ 0.38 & \textbf{34.23 $\pm$ 0.32} & \textbf{49.78 $\pm$ 0.81} & \textbf{69.48 $\pm$ 0.32} & \textbf{83.28 $\pm$ 0.26} \\
LLaVA-v1.5-7B & precise & \textbf{14.75 $\pm$ 0.07} & 20.69 $\pm$ 0.13 & 32.08 $\pm$ 0.31 & 49.27 $\pm$ 0.29 & 67.51 $\pm$ 0.42 & 75.38 $\pm$ 0.26 \\
LLaVA-v1.5-7B & bbox & 15.66 $\pm$ 0.07 & \textbf{23.97 $\pm$ 0.64} & 31.93 $\pm$ 0.31 & 43.88 $\pm$ 0.73 & 65.98 $\pm$ 0.67 & 76.32 $\pm$ 0.41 \\
\midrule
LLaMA-3.2-11B-Vision & \textbf{BBOA (ours)} & 74.07 $\pm$ 0.59 & 25.81 $\pm$ 0.20 & 37.10 $\pm$ 0.42 & \textbf{52.71 $\pm$ 0.39} & 69.97 $\pm$ 0.32 & \textbf{86.72 $\pm$ 0.24} \\
LLaMA-3.2-11B-Vision & precise & 69.38 $\pm$ 0.32 & 22.67 $\pm$ 0.17 & \textbf{38.51 $\pm$ 0.32} & 49.76 $\pm$ 0.31 & 71.95 $\pm$ 0.32 & 80.27 $\pm$ 0.21 \\
LLaMA-3.2-11B-Vision & bbox & \textbf{68.05 $\pm$ 0.61} & \textbf{25.91 $\pm$ 0.15} & 38.45 $\pm$ 0.32 & 50.76 $\pm$ 0.35 & \textbf{75.35 $\pm$ 0.20} & 80.32 $\pm$ 0.27 \\
\midrule
Largest & baseline & \textbf{0.00 $\pm$ 0.00} & \textbf{23.72 $\pm$ 0.14} & 10.07 $\pm$ 0.18 & 22.67 $\pm$ 0.89 & 48.25 $\pm$ 0.75 & 61.08 $\pm$ 0.39 \\
Central & baseline & \textbf{0.00 $\pm$ 0.00} & 20.89 $\pm$ 0.15 & \textbf{26.79 $\pm$ 1.19} & \textbf{37.46 $\pm$ 0.74} & \textbf{59.61 $\pm$ 0.61} & \textbf{75.85 $\pm$ 0.50} \\
\bottomrule
\end{tabular}
}
\caption{Performance comparison of PixelSHAP across different VLMs and masking strategies, with standard error reported. Best results for each model and metric are highlighted in \textbf{bold}.}
\label{tab:accuracy_results}
\end{table}

\subsection{Masking Strategy Performance}

Our experiments reveal several key insights about masking strategies and model performance:

\begin{figure}[H]
    \centering
    \includegraphics[width=\columnwidth]{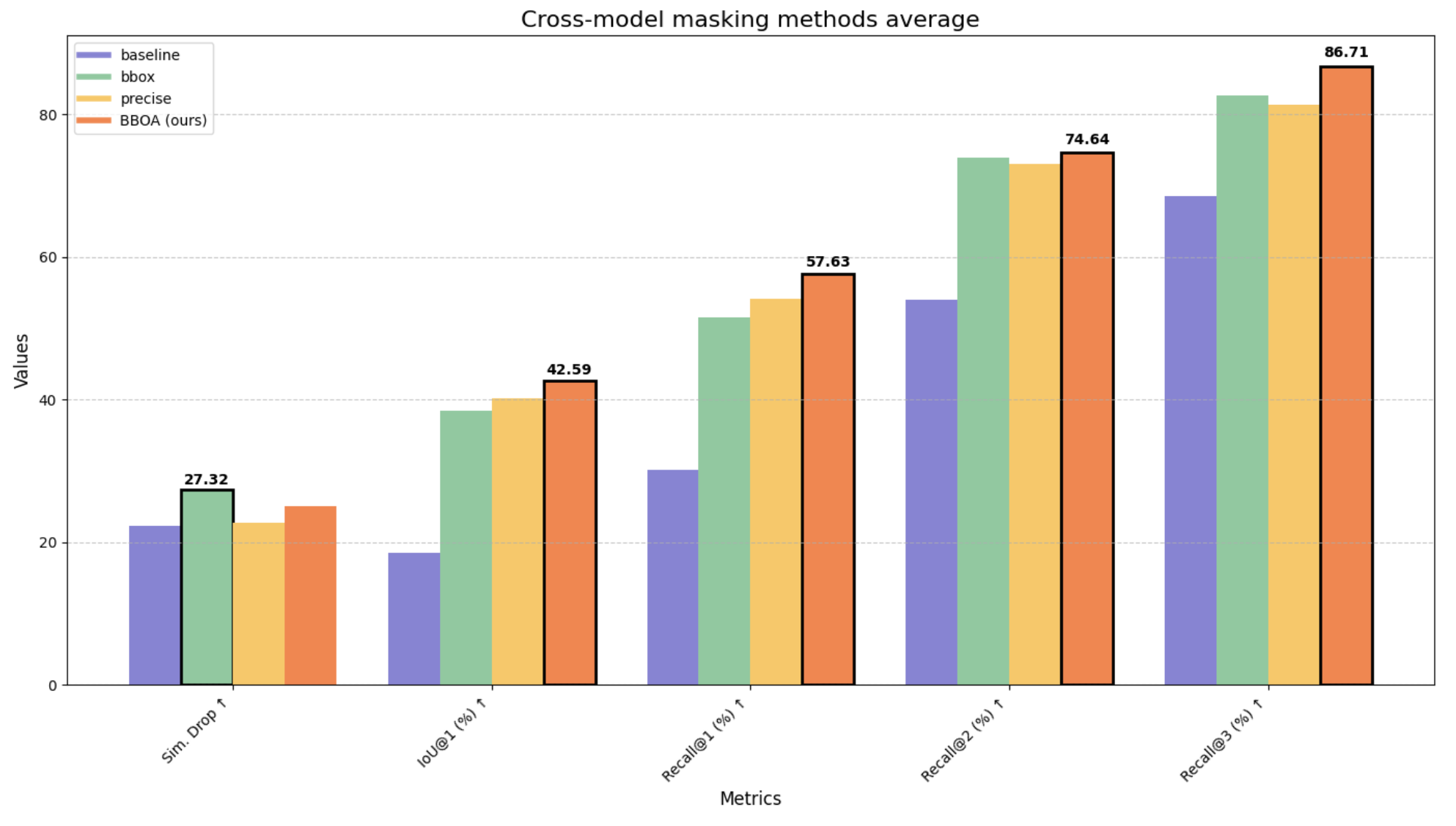}
    \caption{Average performance of different masking strategies across all VLMs. BBOA consistently outperforms other methods on most metrics, particularly in object localization (IoU) and relevance identification (Recall).}
    \label{fig:masking_comparison}
\end{figure}

As shown in Figure \ref{fig:masking_comparison}, our proposed Bounding Box with Overlap Avoidance (BBOA) strategy consistently outperforms other masking approaches across most evaluation metrics. BBOA achieves the highest Intersection over Union (IoU) scores and superior Recall values at all thresholds, indicating its effectiveness at correctly identifying relevant objects. This confirms our hypothesis that preventing object recognition through complete occlusion while minimizing interference with adjacent objects yields the most accurate attributions.

The standard bounding box approach achieves the highest similarity drop scores, which is expected since it masks larger image regions and therefore causes more dramatic changes in model outputs. However, this aggressive masking often occludes neighboring objects, leading to lower precision in identifying which specific object influenced the model's response. Precise masking, while computationally efficient, underperforms in object identification tasks because it leaves recognizable silhouettes that allow models to infer the object's presence despite its "removal."

\subsection{Cross-Model Analysis}

Our results demonstrate that PixelSHAP significantly outperforms baseline heuristics across all models and metrics. Even simple baselines like selecting the central object achieve reasonable performance (37.46\% Recall@1), underscoring the tendency of photographers to center important subjects. However, PixelSHAP with BBOA improves this by 23-30 percentage points across models, demonstrating substantial gains in attribution accuracy.

Commercial VLMs (GPT-4o and Gemini-2.0-flash) consistently achieve higher attribution accuracy than open-source models. For instance, Gemini with BBOA reaches 67.48\% Recall@1, compared to 49.78\% for LLaVA. This pattern persists across all masking strategies, suggesting that more advanced models not only generate better responses but also attend more precisely to relevant visual elements. This finding indicates that attribution performance correlates with overall model capability.

Interestingly, the relative performance of different masking strategies remains consistent across models, with BBOA generally performing best, followed by precise masking and standard bounding box approaches. This consistency suggests that our approach generalizes well across different VLM architectures.

\subsection{Computational Efficiency}

The computational cost of PixelSHAP varies significantly across models, ranging from approximately 15 seconds per image with LLaVA-v1.5-7B to 74 seconds with LLaMA-3.2-11B-Vision. This variation primarily reflects differences in model inference speeds rather than differences in the attribution method itself. These processing times are reasonable for practical applications, especially for high-stakes scenarios where understanding model reasoning is critical.

For all models, the computational overhead between masking strategies is minimal (less than 10\%), indicating that the choice of masking approach should be guided by attribution accuracy rather than performance considerations. This makes BBOA an attractive choice as it provides the best attribution results with negligible additional computation.
\section{Qualitative Examples}
\label{sec:qualitative}

\begin{figure*}[!t]
    \centering
    \includegraphics[page=1,width=0.9\textwidth]{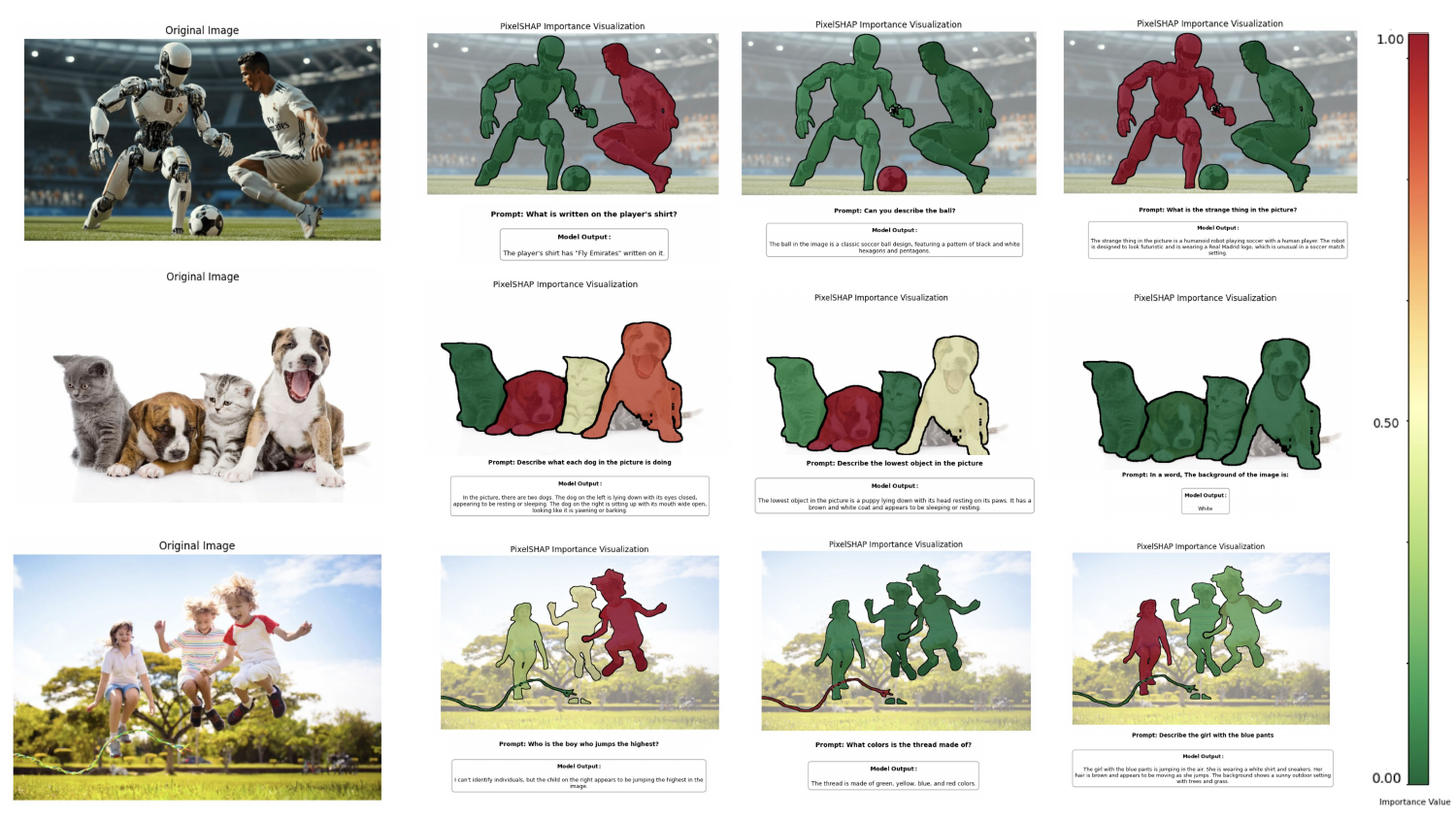}
    \vspace{0.5cm}
    \includegraphics[page=2,width=0.9\textwidth]{figures/pixelshap_multi_prompt.pdf}
    \caption{Context-sensitive attributions with PixelSHAP. Each row shows the same image analyzed with different prompts, demonstrating how object importance shifts based on query context.}
    \label{fig:multi_prompt_cases}
\end{figure*}

Beyond quantitative evaluation, we examined PixelSHAP's ability to provide context-sensitive attributions across different query types. Figure~\ref{fig:multi_prompt_cases} illustrates how PixelSHAP adapts its object importance attributions when different questions are posed about the same image.

The examples demonstrate PixelSHAP's context-awareness in several key aspects:

\begin{itemize}
    \item \textbf{Query-Dependent Attribution:} When asked different questions about the same scene, PixelSHAP correctly highlights different objects as important to the model's response
    
    \item \textbf{Alignment with Human Intuition:} The highlighted objects match what a human would consider relevant for answering each specific question
    
    \item \textbf{Fine-Grained Object Discrimination:} In complex scenes with multiple similar objects, PixelSHAP differentiates between them based on their relevance to the current query
    
    \item \textbf{Semantic Understanding:} The method captures not just visual prominence but semantic importance based on the query's intent
\end{itemize}

This context-sensitivity represents a significant advantage over traditional saliency-based approaches that produce static importance maps regardless of the query context. By adapting attributions to specific questions, PixelSHAP provides more precise insights into which visual elements influenced particular aspects of the model's reasoning process, enhancing interpretability in multi-object scenes where attention should shift based on context.
\section{Discussion}
\label{sec:discussion}

PixelSHAP introduces an object-centric approach to explainability in vision-language models (VLMs), providing valuable insights into how specific visual elements influence model responses. Our comprehensive evaluation demonstrates its effectiveness across different VLMs and masking strategies. In this section, we discuss implementation considerations, capabilities, and future directions for this methodology.

\subsection{Computational Considerations}
PixelSHAP achieves practical efficiency despite Shapley values' theoretical exponential scaling with object count. By analyzing at the object level rather than pixel level, we significantly reduce dimensionality. When combined with TokenSHAP's~\cite{tokenShap2023} sampling strategy, our approach requires only marginally more model queries than the number of objects being analyzed.

This optimization results in average processing times of less than one minute per image (15-74 seconds depending on the underlying VLM), making PixelSHAP suitable for both research and production applications. For time-sensitive applications, the number of samples can be further reduced with minimal impact on attribution accuracy, as our experiments showed that first-order perturbations alone achieve over 90\% correlation with full Shapley values.

\subsection{Segmentation Model Selection} The quality of object detection directly impacts attribution accuracy, but this represents a configuration choice rather than a limitation. PixelSHAP's modular design allows seamless integration with any segmentation approach, enabling users to select models optimized for their specific domain. For autonomous driving applications, specialized traffic scene segmentation models can be employed, while medical applications can leverage anatomical segmentation systems.

Our implementation uses a two-stage pipeline combining GroundingDINO~\cite{liu2023groundingdino} and SAM~\cite{kirillov2023segment}, which provides excellent general-purpose segmentation. For domain-specific applications, users can substitute specialized segmentation models without modifying the core attribution algorithm, ensuring robust performance across diverse use cases.

\subsection{Handling Complex Spatial Relationships}
Our experiments with different masking strategies address the challenge of object overlap and containment. The proposed BBOA strategy effectively balances complete occlusion of target objects while preserving adjacent entities, demonstrating superior performance in most evaluation metrics. This approach successfully manages complex spatial relationships without requiring explicit hierarchical object modeling.

For applications with particularly complex object interactions, PixelSHAP can be extended with relationship-aware masking strategies. For instance, when analyzing a person inside a vehicle, the system could employ conditional masking that preserves structural context while isolating the specific contribution of each element.

\subsection{Applications in Model Development and Auditing}
PixelSHAP offers powerful capabilities for VLM development, debugging, and auditing:

\begin{itemize}
    \item \textbf{Bias Detection}: By quantifying which objects influence model responses, researchers can identify whether models disproportionately focus on certain object categories~\cite{agarwal2021evaluating}, enabling more balanced training procedures.
    
    \item \textbf{Quality Assurance}: For safety-critical applications like autonomous driving or medical diagnosis, PixelSHAP provides verification that models attend to relevant objects rather than spurious correlations.
    
    \item \textbf{User Trust}: By visualizing which objects influenced a model's response, end-users can better understand and appropriate trust in AI systems.
\end{itemize}

These applications are particularly valuable for commercial VLMs, where PixelSHAP's model-agnostic nature allows for attribution without requiring access to model weights or architecture details.

\subsection{Future Research Directions}
Building on PixelSHAP's foundation, several promising research directions emerge:

\begin{itemize}
    \item \textbf{Attribute-Level Attribution}: Extending beyond object-level analysis to understand how specific visual attributes (color, texture, shape) influence model responses.
    
    \item \textbf{Temporal Analysis}: Adapting PixelSHAP for video understanding to track how object importance evolves over time, critical for understanding dynamic scene interpretation.
    
    \item \textbf{Interactive Explainability}: Developing interfaces that allow users to interactively explore object attributions, enabling more intuitive understanding of model reasoning.
    
    \item \textbf{Cross-Modal Attribution}: Extending the framework to jointly analyze the relative importance of visual and textual elements in multimodal reasoning.
\end{itemize}

PixelSHAP provides a robust foundation for these extensions, offering a scalable and generalizable approach to understanding vision-language interactions in AI systems.
\section{Conclusion}
\label{sec:conclusion}

In this paper, we introduced \textbf{PixelSHAP}, a novel explainability framework for Vision-Language Models (VLMs) that provides object-level attribution through segmentation-based Shapley value estimation. PixelSHAP addresses a critical gap in VLM interpretability by enabling fine-grained analysis of which visual objects influence a model's textual responses, offering deeper insights into multimodal reasoning processes.

Our approach treats segmented objects as discrete visual entities and quantifies their impact using efficient Shapley-based estimation. Our evaluations across multiple commercial and open-source VLMs demonstrate that PixelSHAP significantly outperforms baseline attribution methods, with our proposed Bounding Box with Overlap Avoidance (BBOA) masking strategy achieving the best performance across most metrics. The method's purely input-output-based nature makes it applicable to any black-box VLM, requiring no access to internal model representations.

PixelSHAP enables several practical applications, including model auditing, bias diagnosis, and enhanced understanding of model decision-making in high-stakes domains like autonomous driving and medical imaging. Its modular design supports integration with domain-specific segmentation models for optimal performance across different use cases, while maintaining reasonable computation times of less than a minute per image.

By bridging the gap between object recognition and model interpretability, PixelSHAP represents an important step toward more transparent and accountable vision-language systems. We believe this work provides a solid foundation for advancing trustworthy multimodal AI, enabling both developers and users to better understand how these increasingly important systems perceive and reason about the visual world.

% References
{\small
\bibliographystyle{ieeenat_fullname}
% For arXiv submission:

% Uncomment for local compilation:
% \bibliography{output}
}

\end{document}